\documentclass[10pt,twocolumn,letterpaper]{article}

\def\cvprPaperID{0000} 
\usepackage{cvpr}
\usepackage{times}
\usepackage{epsfig}
\usepackage{graphicx}
\usepackage{caption}
\usepackage{subcaption}
\usepackage{float}
 
\usepackage{amsmath}
\usepackage{amssymb}

\usepackage{xcolor}
\usepackage{xurl}
\usepackage{enumitem}
\setlist[itemize]{leftmargin=*} 
\setlist[enumerate]{leftmargin=*}

\usepackage{multirow}
\usepackage{booktabs}
\usepackage{environ}
\usepackage{mdframed}

\usepackage{todonotes}

\usepackage{lipsum}

\usepackage[pagebackref=true,breaklinks=true,letterpaper=true,colorlinks,bookmarks=false]{hyperref}
\hypersetup{colorlinks,linkcolor={black},citecolor={black},urlcolor={black}} 

\cvprfinalcopy 

\def\cvprPaperID{****} 

\ifcvprfinal\fi
\begin{document}

\title{Towards a Unified Multimodal Reasoning Framework 
}

\author{Abhinav Arun\\
{\tt\small aarun60@gatech.edu}\\
\and
Dipendra Singh Mal\\
{\tt\small dmal3@gatech.edu}\\
\and
Mehul Soni\\
{\tt\small mehul.soni918@gatech.edu}\
\and
Tomohiro Sawada\\
{\tt\small tsawada7@gatech.edu}\\
}

\maketitle

\begin{abstract}

Recent advancements in deep learning have led to the development of powerful language models (LMs) that excel in various tasks. Despite these achievements, there is still room for improvement, particularly in enhancing reasoning abilities and incorporating multimodal data. This report investigates the potential impact of combining Chain-of-Thought (CoT) reasoning and Visual Question Answering (VQA) techniques to improve LMs' accuracy in solving multiple-choice questions. By employing TextVQA and ScienceQA datasets, we assessed the effectiveness of three text embedding methods and three visual embedding approaches. Our experiments aimed to fill the gap in current research by investigating the combined impact of CoT and VQA, contributing to the understanding of how these techniques can improve the reasoning capabilities of state-of-the-art models like GPT-4. Results from our experiments demonstrated the potential of these approaches in enhancing LMs' reasoning and question-answering capabilities, providing insights for further research and development in the field, and paving the way for more accurate and reliable AI systems that can handle complex reasoning tasks across multiple modalities.
\end{abstract}

\section{Introduction/Background/Motivation}





The advent of deep learning has led to the development of powerful language models (LMs) that excel in various tasks, such as question-answering\cite{gupta2023instruction}, translation \cite{Lewis2019BARTDS}, sentiment analysis \cite{scaria2023instructabsa}, event detection \cite{anantheswaran2023edm3} and synthetic data creation \cite{gupta2023targen}. However, there is still potential for improvement, particularly in enhancing their reasoning abilities and incorporating multimodal data like images. This report investigates the potential impact of combining Chain-of-Thought reasoning (CoT) \cite{Wei22,gupta2022john} and Visual Question Answering (VQA) \cite{Singh19} techniques to improve LMs' accuracy in solving multiple-choice questions.

The primary objective of this research is to explore the synergistic effects of CoT and VQA on LMs' performance. CoT involves generating rationales for each choice, providing a logical explanation for the model's decision-making process. VQA includes using images as additional information to answer questions. By combining these techniques, we aim to demonstrate substantial improvements in LMs' reasoning and question-answering capabilities.

To assess these techniques' effectiveness, we experimented with three text embedding methods and three visual embedding approaches. Current research primarily focuses on CoT and VQA individually. Our project aims to fill the gap by investigating the combined impact of CoT and VQA, contributing to the understanding of how these techniques can improve the reasoning capabilities of state-of-the-art models like GPT-4\cite{openai2023gpt4}.

The datasets utilized in our experiments are TextVQA \cite{TextVQA} and ScienceQA \cite{ScienceQA}. TextVQA consists of 45,336 questions based on 28,408 images from the Open Images dataset, requiring reasoning and reading about text in the image. ScienceQA contains 21,208 multimodal multiple-choice science questions collected from elementary and high school science curricula, covering a wide range of subjects and offering a rich diversity of domains.

By employing these datasets and combining CoT and VQA techniques, we strive to demonstrate the potential of these approaches in improving LMs' reasoning and question-answering capabilities. The successful integration of these techniques will contribute to the understanding of deep learning models' underlying mechanisms, ultimately leading to the development of more efficient and accurate LMs for a wide range of applications.

\begin{figure*}[t!]
    \begin{minipage}[b]{\textwidth}
            \begin{tabular}{|l|l|l|l|l|l|l|}
            \hline
            \textbf{Model} & \textbf{Generation Task} & \textbf{Train Accuracy} & \textbf{Validation Accuracy} & \textbf{Rouge-1 F1} & \textbf{Rouge-2 F1} & \textbf{Rouge-L F1} \\ \hline
            Model 1 & Answer & - & 48.74\% & - & - & - \\ \hline
            Model 2 & Answer & 45.66\% & 43.65\% & - & - & - \\ \hline
            Model 3 & Answer, Explanation & 41.32\% & 39.09\% & 0.41 & 0.298 & 0.375 \\ \hline
            Model 4 & Answer, Explanation & 48.54\% & 45.84\% & 0.451 & 0.331 & 0.408 \\ \hline
            Model 5 & Answer (Input Explanation) & 45.01\% & 41.99\% & - & - & - \\ \hline
            Model 6 & Answer (Input Explanation) & 51.76\% & 48.86\% & - & - & - \\ \hline
            \end{tabular}
            \caption{Evaluation metrics for models on the generation downstream tasks on ScienceQA Dataset}
            \hbox{\label{sec:model1}\textbf{Model 1}: Baseline pre-trained T5 model fine-tuned for the task of answer generation without the usage of image captions.}
            \hbox{\label{sec:model2}\textbf{Model 2}: Baseline pre-trained T5 model fine-tuned for the task of answer generation with the usage of image captions.}
            \hbox{\label{sec:model3}\textbf{Model 3}: Baseline pre-trained T5 model fine-tuned for the task of answer and explanation generation with the image captions.}
            \hbox{\label{sec:model4}\textbf{Model 4}: Model fine-tuned for the task of generating both answers and explanations using the fine-tuned model for answer generation} 
            \hbox{\label{sec:model5}\textbf{Model 5}: Baseline pre-trained T5 model fine-tuned for the task of answer generation with the model generated explanation as input}
            \hbox{\label{sec:model6}\textbf{Model 6}: fine-tuned Model 2 for the task of answer generation with the model-generated explanation as input}
            \label{tab:main_results}
    \end{minipage}
    \vspace{1em}
    \begin{minipage}[b]{.32\textwidth}
    \includegraphics[width=\linewidth]{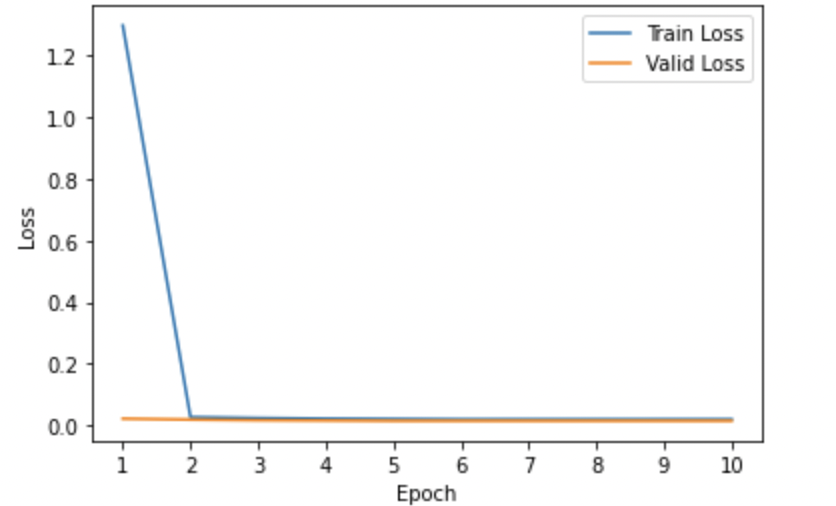}
    \caption{Model 1}
    \label{fig:model8}
    \end{minipage}
    \hspace{\fill} 
    \begin{minipage}[b]{.32\textwidth}
    \includegraphics[width=\linewidth]{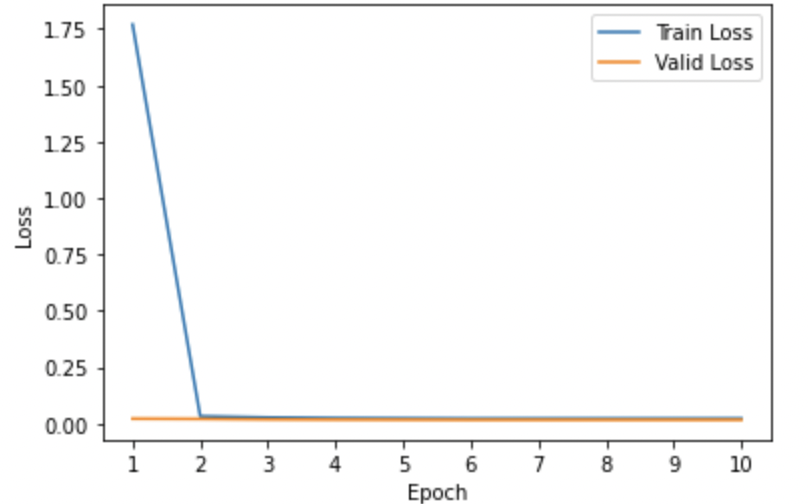}
    \caption{Model 2}
    \label{fig:model2}
    \end{minipage}
    \hspace{\fill} 
    \begin{minipage}[b]{.32\textwidth}
    \includegraphics[width=\linewidth]{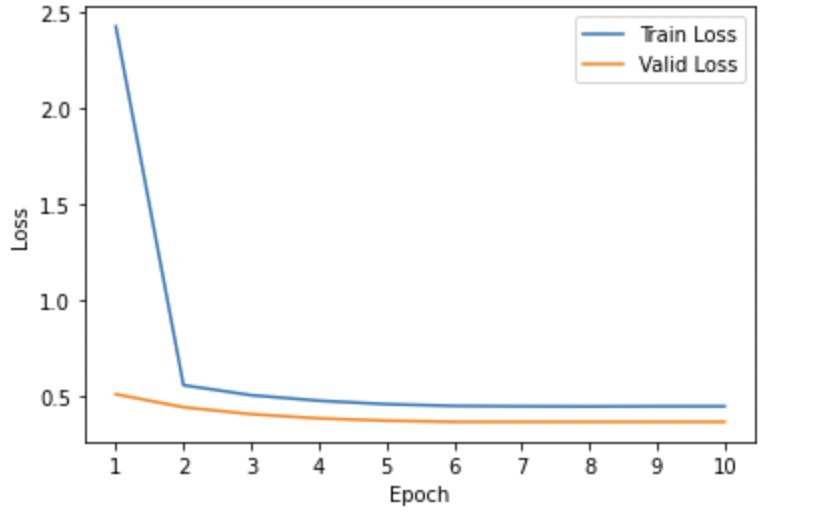}
    \caption{Model 3}
    \label{fig:model3}
    \end{minipage}
    
    \vspace*{1cm} 
    
    \begin{minipage}[b]{.32\textwidth}
    \includegraphics[width=\linewidth]{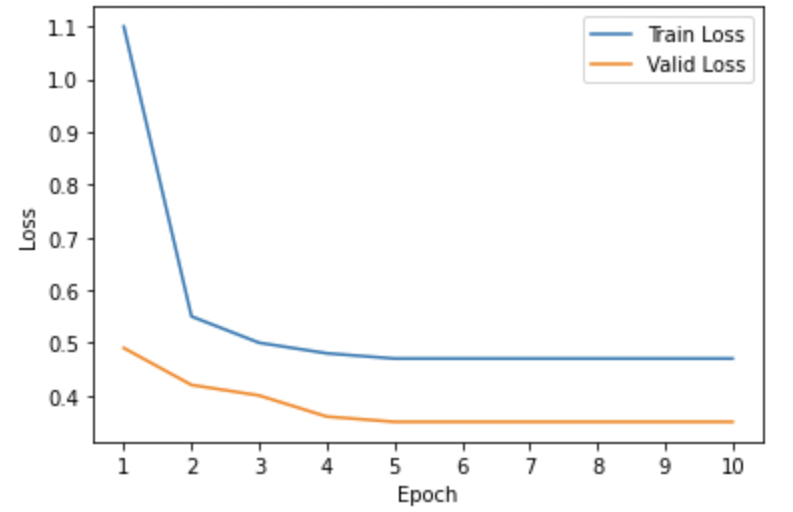}
    \caption{Model 4}
    \label{fig:model6}
    \end{minipage}
    \hspace{\fill}
    \begin{minipage}[b]{.32\textwidth}
    \includegraphics[width=\linewidth]{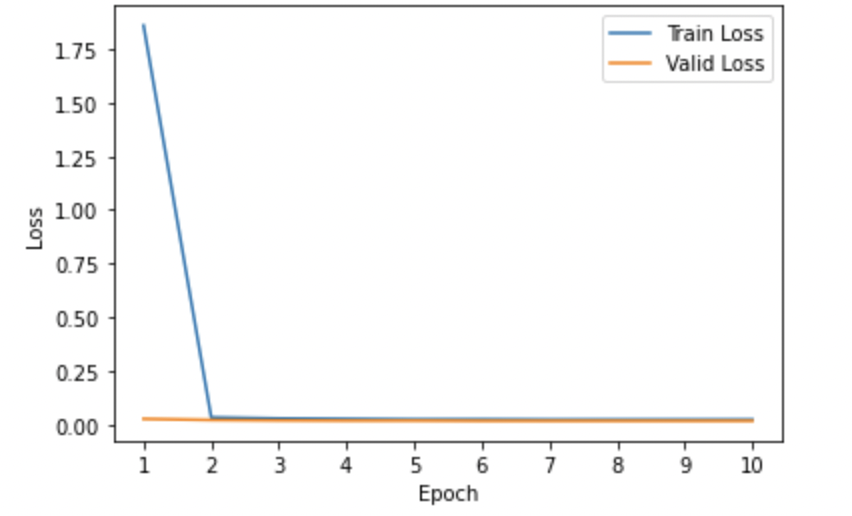}
    \caption{Model 5}
    \label{fig:model5}
      \end{minipage}
      \hspace{\fill}
    \begin{minipage}[b]{.32\textwidth}
    \includegraphics[width=\linewidth]{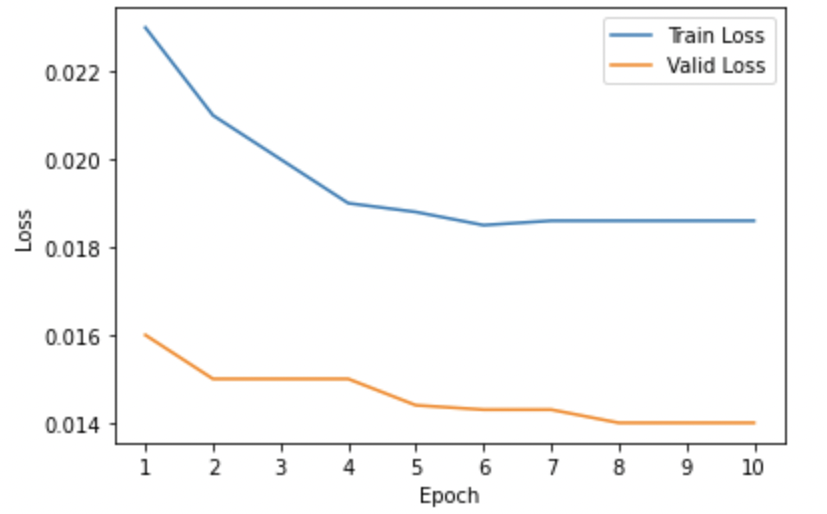}
    \caption{Model 6}
    \label{fig:model14}
   \end{minipage}
\end{figure*}

\section{Approach}




The primary objective of this project is to explore the synergistic effects of CoT and VQA on LMs' performance. CoT involves generating rationales for each choice, providing a logical explanation for the model's decision-making process \cite{Wei22}. VQA includes using images as additional information to answer questions \cite{Singh19}. By combining these techniques, we aim to demonstrate substantial improvements in LMs' reasoning and question-answering capabilities.

To assess these techniques' effectiveness, we experimented with three text embedding methods and three visual embedding approaches. Current research primarily focuses on CoT and VQA individually \cite{Wei22, Singh19}. Our project aims to fill the gap by investigating the combined impact of CoT and VQA, contributing to the understanding of how these techniques can improve the reasoning capabilities of state-of-the-art models like GPT-4 \cite{SOTA23}.

The datasets utilized in our experiments are TextVQA and ScienceQA \cite{TextVQA, ScienceQA}. TextVQA consists of 45,336 questions based on 28,408 images from the Open Images dataset, requiring reasoning and reading about text in the image \cite{Singh19}. ScienceQA contains 21,208 multimodal multiple-choice science questions collected from elementary and high school science curricula, covering a wide range of subjects and offering a rich diversity of domains \cite{Geva2021DidAU}

We set up the following text embedding models for our experiments: 1) a simple QA model using DistilBERT as a baseline \cite{Sanh19}; 2) T5 with reasoning without image captions \cite{Khashabi20}; and 3) T5 with reasoning and image captions \cite{Khashabi20}. For visual embedding models, we considered: 1) a baseline VQA model \cite{Li19}; 2) integrating visual embeddings with textual embeddings for the baseline textual model \cite{Lu19}; and 3) visual embedding with textual embeddings for the T5 model \cite{Kim21}.

We thought this would be a fruitful approach since VQA and CoT individually already improved model performance substantially on similar benchmarks \cite{Wei22, Singh19}. Using VQA together with CoT is a new approach, which we aimed to explore in this study \cite{SOTA23}.

Anticipated problems included limitations with computational resources, and we encountered some models returning surprisingly poor performance (RoBERTa) contrary to expectations \cite{Liu19}. The very first thing we tried did not work, but we iteratively refined our approach to address these issues.\\

We did not use any code from repositories, but we used the following for reference:

\begin{enumerate}
    \item Document Question Answering \cite{HuggingFaceDocsDQA}
    \item Question Answering With T5 | Kaggle \cite{abdokamr_2021}
    \item Towards Data Science: Adding Custom Layers on Top of a Hugging Face Model \cite{TDS20}
    \item Multiple choice \cite{MultipleChoice}
\end{enumerate}

By combining CoT and VQA techniques, we strive to demonstrate the potential of these approaches in enhancing the reasoning and question-answering capabilities of LMs. Our experiments may pave the way for further research and development in the field, leading to more accurate and reliable AI systems that can handle complex reasoning tasks across multiple modalities.

By integrating CoT and VQA, we hope to leverage the strengths of both approaches, enabling LMs to reason more systematically and accurately when provided with textual and visual information. This combined approach could be particularly useful for real-world applications where data comes in various forms and requires the integration of multiple sources of information for effective decision-making.

Moreover, our research could inspire future work on novel architectures and algorithms that capitalize on the synergies between CoT and VQA, pushing the boundaries of what is possible with current AI technology. Our findings may also contribute to the development of more interpretable models, as generating rationales for each choice can help explain the model's decision-making process, making it more transparent and understandable for human users.\\

Link to our code implementations : \href{https://github.com/tomohiro-sawada/cs7643-final-project}{\textbf{CS 7643 Project Code}}

\section{Experiments and Results}



To evaluate the success of our proposed approach, we conducted a series of experiments using various models and configurations. We then compared the results, both quantitatively and qualitatively, to assess the effectiveness of our approach in enhancing the reasoning and question-answering capabilities of LMs.

The memory and computational requirements needed for creating scalable VQA models constrainted us with using models that use both image and text features for question answering. We tried a couple of multi-modal frameworks like ViLT \cite{huggingface2021vilt} and VisualBERT \cite{huggingface2021visualbert}. We fine-tuned the ViLT model on the ScienceQA dataset by manually creating domain-specific vocabulary and annotations with scores for probable answers. The model, however, did not perform well on the dataset as it was constrained to generating single-word answers and lacked the capability of generating coherent reasoning like Text-to-Text Language models (T5 \cite{huggingface2020t5}).


\subsection{TextQA Tasks}

We started with a textual question-answering task to evaluate the reasoning capabilities of our models. The following models were used in our experiments:
\begin{itemize}

\item Baseline: A simple QA model based on DistilBERT and RoBERTa. These models were chosen due to their relatively small size, making them suitable for training with our computational resources. Moreover, they have been shown to perform decently on multiple-choice questions.
  
\item T5 without reasoning: We trained a T5-small model without any reasoning capabilities to assess the impact of adding CoT reasoning.
  
\item T5 with reasoning: We further trained a T5-small model with reasoning capabilities, integrating the CoT approach.
  
\item T5 with reasoning and image captions: To assess the potential benefits of adding image information, we trained a T5-small model with reasoning capabilities and image captions.

\end{itemize}

\subsection{VisualQA Tasks}

In the second phase of our experiments, we focused on the VQA task, integrating visual embeddings with textual embeddings.
\begin{itemize}

\item Fine-tuned VQA model on Science QA Dataset: We fine-tuned pre-trained ViLT(Vision \& Language Transformer) model for visual question answering. We created a domain-specific vocabulary and annotations using the ScienceQA dataset and used the ViLT model to generate answers. The ViLT model, however, did not perform well on the dataset as it was constrained to generating single-word answers and lacked the capability of generating coherent reasoning like Text-to-Text Language models (T5).

\item We also attempted to integrate the visual embeddings from models like DETR into the VisualBert model. However, we were not successfully able to concatenate the visual and text features to re-train the VisualBert model. This was due to the varying hidden dimension of the textual encodings for different downstream task models in VisualBert.
  
\item Integrated model with T5 textual embeddings: We generated image captions from a Vision Transformer model (ViT-GPT2 \cite{huggingface2021vit}) and  used that along with textual input to T5-small model and experimented with different training strategies and settings to assess the impact of adding reasoning capabilities and image context. 

We ran our experiments for answer generation on both TextVQA and ScienceQA dataset along with setting up the training for Answer and Explanation generation solely on the ScienceQA dataset where we had ground truth explanations (solutions) for which we measured the Rogue F1 scores.
\end{itemize}

\subsection{Experimental Settings and Hyperparameters}
We set-up our experiments on Google Cloud Platform using GPU setting for the training and evaluation of our models. For the experiments with the T5 model, we used the Adam Optimizer for training with a learning rate of 1e-5. We also used a linear learning rate scheduler with number of training steps as 10000 and a number of warm up steps as 500. We set the max input and output length depending on the downstream task we attempted to solve (128 for answer generation and 256 for explanation generation). 

We also explored gradient clipping to avoiding exploding gradient in the training strategy. We monitored both the training and validation loss after each epoch with the total number of epochs finally set to 10.
\subsection{Metrics}

To measure the performance of the models, we used the following metrics:

\begin{enumerate}
    \item Accuracy: We computed the accuracy by comparing the model's predictions with the ground truth answers.
    \item ROUGE $F_{1}$ score: These metrics evaluate the quality of the generated text by comparing it to the reference text. They provide a quantitative measure of the model's performance. The $F_{1}$ score was chosen as it provides a balance between the precision and recall.
\end{enumerate}

\begin{figure}[H]
  \begin{minipage}[b]{0.5\textwidth}
  \centering
    \includegraphics[scale=0.21]{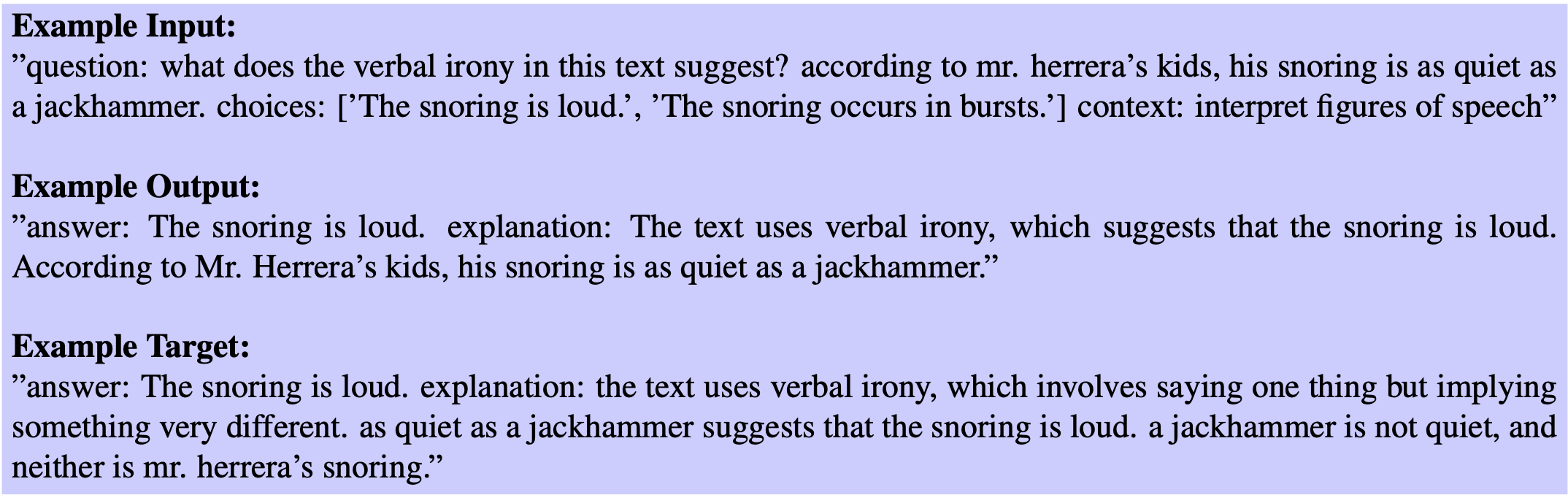}
    \caption{This image shows the T5 model output generating answer and explanation simultaneously. It also has the example input (which is a mix of question, choices, context, and hint) and the target output which the model uses in its training phase}
    \label{}
  \end{minipage}
\hfill
    \begin{minipage}[b]{0.5\textwidth}
  \centering
    \includegraphics[scale=0.21]{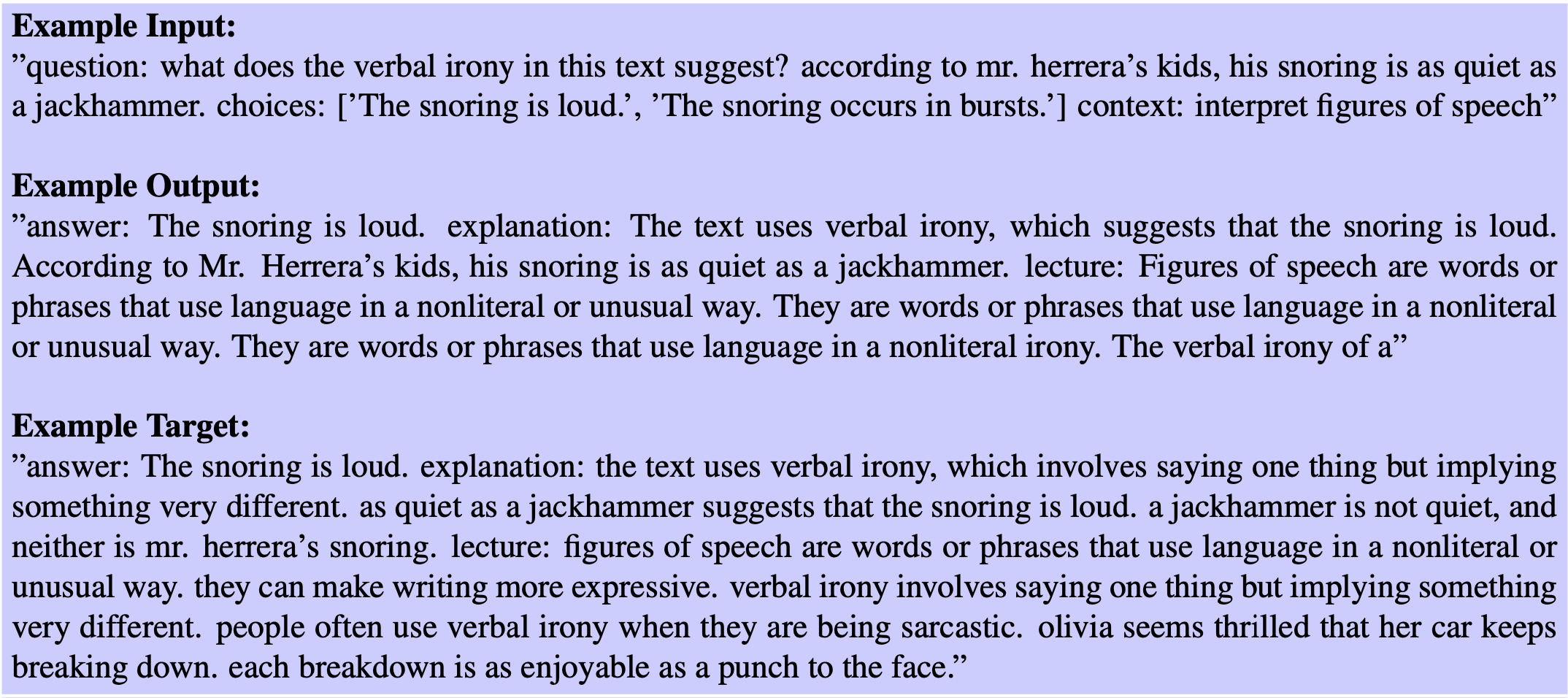}
    \caption{This image shows the same T5 model as in Figure 7 finetuned for the task of generating answer, explanation, and lecture simultaneously. This produces a rich expression and reasoning behind the model's answers}
    \label{}
  \end{minipage}
  \hfill
  \begin{minipage}[b]{.5\textwidth}
  \centering
    \includegraphics[scale=0.21]{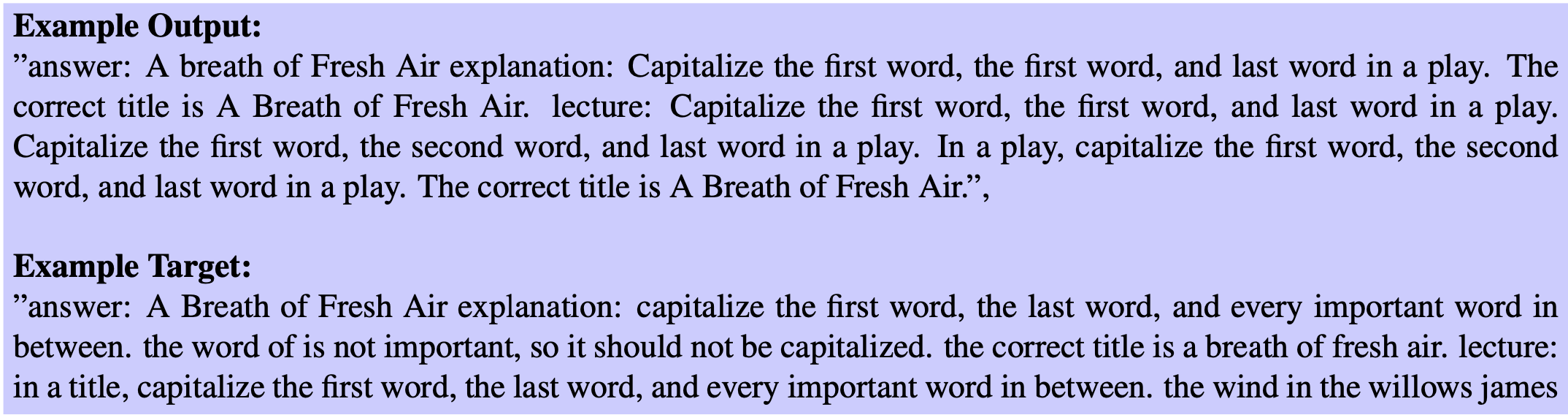}
    \caption{This is an example of hallucination. This explains the need for teacher training or feedback mechanism - i.e. using the model-generated explanations to infer the answers. A lot of times we encounter such hallucination examples where the answer generated by the model is at odds with the model-generated example. Thus, in these cases , 2-stage framework works best which incorporates the generated explanation to produce answers in the subsequent run}
    \label{}
  \end{minipage}
\end{figure}

\begin{figure}[H]
    \centering 
    \includegraphics[scale=0.6]{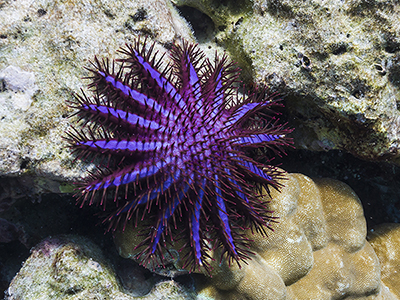}
    \hbox{\textbf{Questions}: Which is this organism's common name?}
    \hbox{\textbf{Choices}: crown-of-thorns sea star, Acanthaster planci}
    \hbox{\textbf{Caption}: a blue and white vase sitting on top of a rock}
    \hbox{\textbf{T5 model's answer}: crown-of-thorns sea star}
    \hbox{\textbf{ViLT Model's Answer}: fish}
    \hbox{\textbf{Target Answer}: crown-of-thorns sea star}
    \label{}
    \caption{This image clearly explains why the T5 model is pretty good in answering domain-specific questions after properly finetuning the model. The ViLT model for question answering produces a very generic response and is not able to output coherent sentences as it generates a probability distribution over the entire vocabulary.}
\end{figure}

\subsection{Key Findings}

\begin{itemize}
    \item Training and Validation Loss Curves: We analyzed the loss curves specific to all 6 model variants illustrated using Figures 1-6 respectively. Most of the curves showed a sharp dip in the training loss in the first epoch suggesting that the pre-trained T5 model was pretty quick in re-adjusting its weights to better align with the domain-specific examples. The validation losses in most cases were pretty low from the initial get-go. Also, we observed convergence in the training and validation losses for almost all model variants suggesting that running the model for more than 10 epochs would pose the risk of running into an overfitting problem. Also , for (\hyperref{sec:model6}{Model6}) which we found to be the best-performing model, we could see pretty low values of both training and validation losses when compared to other models from initial epochs.
    \item The models that were trained using image captions along with context and hint 
    (\hyperref[sec:model2]{Model 2}) 
    did not outperform the model without the image captions (\hyperref[sec:model1]{Model 1}) when compared on the validation dataset. This was slightly counter-intuitive but the reason for it might be that the model suffered from information overload and could not specifically utilize the captions when provided along with hint and context. This also highlights the importance of working with models that can exploit the mutual synergies of different modalities like text and vision features with an attention mechanism to generate coherent reasoning. The idea of using the image features as textual image captions did not yield enhanced performance.
    \item The model that was trained to generate answers and explanations simultaneously (\hyperref[sec:model3]{Model3}) was outperformed by the model trained on just generating answers (\hyperref[sec:model2]{Model 2}) . 
    This emphasizes the fact that task-specific training gives  better results and prevents the model from information load. 
    \item The models that were directly fine-tuned from the pre-trained T5 conditional model on generating both answers and explanations simultaneously (\hyperref[sec:model3]{Model3}) was outperformed by the model (\hyperref[sec:model4]{Model4})  which was fine-tuned first on generating answers and then using that model checkpoint to generate answers and explanations. This makes sense as it emphasizes that the fine-tuned model on a domain-specific dataset could leverage its previous learnings to perform well on subsequent runs.
    \item Knowledge Distillation with teacher training: We found that the models which utilized the generated explanations from the same model in the previous run (\hyperref[sec:model5]{Model5}, \hyperref[sec:model6]{Model 6}) outperformed the model trained on directly generating answers respectively 
    (\hyperref[sec:model2]{Model 2})  
    This is also intuitive as using the model-generated explanations acts as positive feedback as the model learns to better its prediction on the answers using that as additional input.
    \item Overall, for the downstream task of answer generation, the best result was observed for the model (\hyperref{sec:model6}{Model6}) that was fine-tuned using the previously trained model on answer generation using the generated explanation in the previous run as input (feedback loop - teacher training).
    \item For the downstream task of answer generation, we also see that the T5-small model trained on the TextVQA dataset outperformed the other models tested on the ScienceQA dataset with a training accuracy of 65.92\% and an accuracy of 62.66\%. This is in accordance with our hypothesis as this dataset had more training examples and all the examples contained images unlike the ScienceQA dataset. 
    Additionally, a lot of the questions in the ScienceQA dataset involved very domain-specific vocabulary and trickier questions for the model to learn from given only the caption and hint which is why we wish to further explore models that take into account vision features as well.
\end{itemize}

\pagebreak
{\small
\bibliographystyle{ieee_fullname}
\bibliography{cvpr}
}

\end{document}